\documentclass[sigconf,nonacm]{acmart}
\usepackage{booktabs, makecell, multirow, paralist}

\AtBeginDocument{%
  \providecommand\BibTeX{{%
    \normalfont B\kern-0.5em{\scshape i\kern-0.25em b}\kern-0.8em\TeX}}}

\begin{document}

\title{Heterogeneous Line Graph Transformer for Math Word Problems}

\author{Zijian Hu}
\affiliation{%
  \institution{Southeast University}
  \city{Nanjing}
  \country{China}}
\email{zijian@seu.edu.cn}

\author{Meng Jiang}
\affiliation{%
  \institution{University of Notre Dame}
  \city{Notre Dame}
  \state{Indiana}
  \country{USA}}
\email{mjiang2@nd.edu}

\begin{abstract}
This paper describes the design and implementation of a new machine learning model for online learning systems. We aim at improving the intelligent level of the systems by enabling an automated math word problem solver which can support a wide range of functions such as homework correction, difficulty estimation, and priority recommendation.
We originally planned to employ existing models but realized that they processed a math word problem as a sequence or a homogeneous graph of tokens. Relationships between the multiple types of tokens such as entity, unit, rate, and number were ignored.
We decided to design and implement a novel model to use such relational data to bridge the information gap between human-readable language and machine-understandable logical form.
We propose a heterogeneous line graph transformer (HLGT) model that constructs a heterogeneous line graph via semantic role labeling on math word problems and then perform node representation learning aware of edge types.
We add numerical comparison as an auxiliary task to improve model training for real-world use.
Experimental results show that the proposed model achieves a better performance than existing models and suggest that it is still far below human performance. Information utilization and knowledge discovery is continuously needed to improve the online learning systems.
\end{abstract}

\maketitle

\section{Introduction}

Online learning systems have shown rapid growth in recent years, such as coursera.org, KhanAcademy.org and study.163.com. Statistics from Coursera indicate that over 77 million users are currently learning on the platform. With this trend comes an increasing demand for the intelligence level of online education systems \cite{LiuHHLCSH18, QiuW019, AndersonHKL14}. Thousands of learning functions have been developed on these systems to satisfy personalized needs.

\begin{figure}[t]
  \centering
  \includegraphics[width=\linewidth]{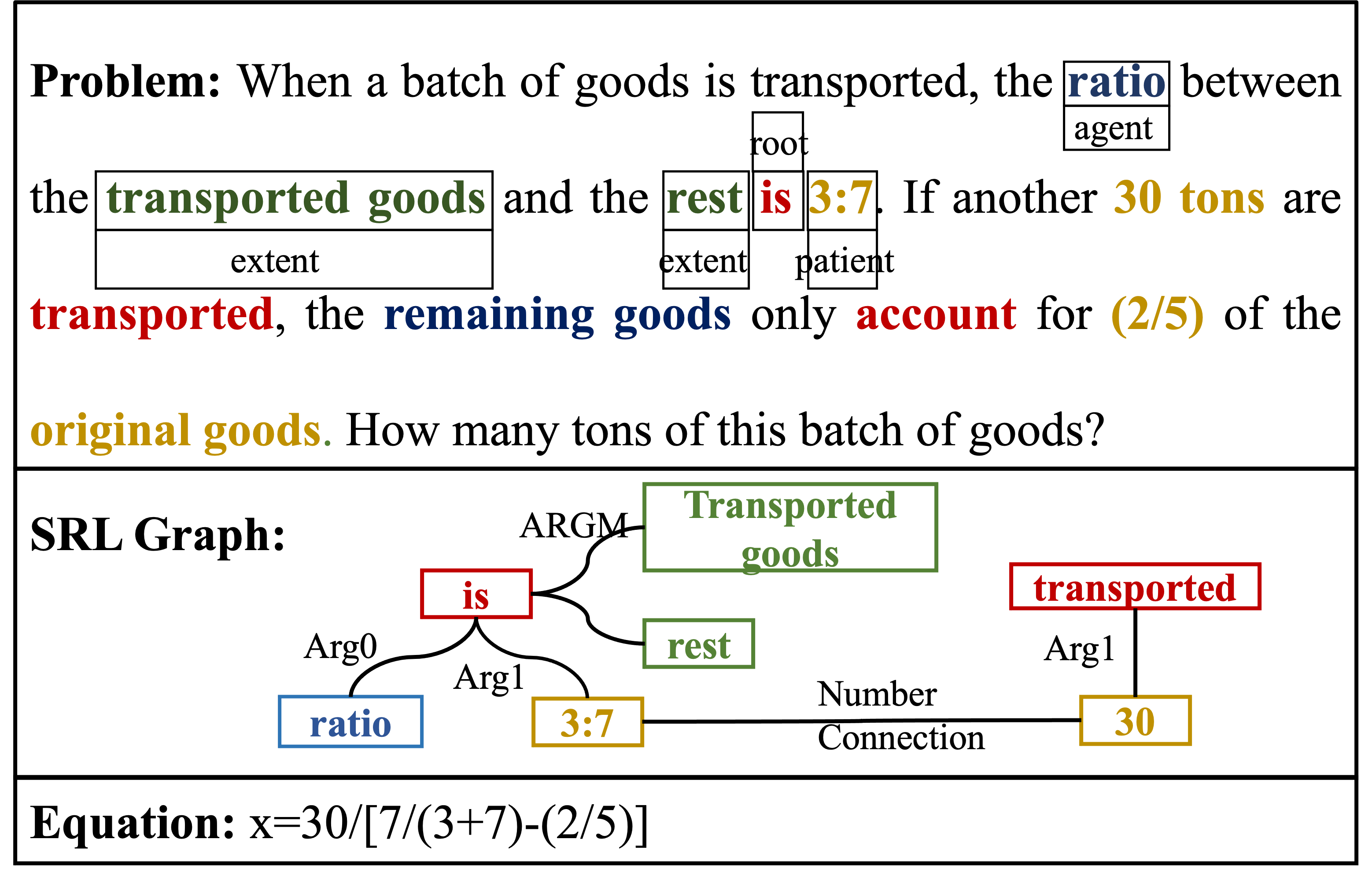}
  \vspace{-0.2in}
  \caption{A math word problem (MWP) example with its semantic role labeling (SRL) graph: Given the question, we aim at producing the equation and answer automatically.}
  \label{fig:1}
  \vspace{-0.2in}
\end{figure}

Building automated math word problem (MWP) solver provides a significant test for the intellectual abilities of these applications \cite{HuangSLY17}. On one hand, it evaluates whether the problem representations in online learning systems could understand and work with numbers \cite{LinHZCLWW21, YuDLZ00L18, DuaWDSS019}. On the other hand, it brings new possibilities for downstream applications, e.g., answering new problems that were not seen in the database \cite{RajpurkarJL18}. However, the performance of the current MWP solver is far from providing sufficient support for different question-based applications. That is mainly because the learned representations of educational questions is not informative enough \cite{ZhangWLBWSL20, ShenJ20}. It is appealing to provide a MWP representation enhancement module to improve the performance of the systems.


As a fundamental issue to test the intelligence of online education systems, automated MWP solver has attracted lots of research interest. At the initial stage, statistical machine learning and semantic parsing methods were mainstream solutions \cite{KushmanZBA14, HosseiniHEK14, RoyR18, ShiWLLR15, Koncel-Kedziorski15}, with the objective of mapping the sentences of problem statements into structured logic representations so as to facilitate quantitative reasoning. Recently, with the availability of large-scale training data, people attempted to exploit the expressive power of deep learning models. The deep learning models achieved better performance than previous methods because it avoided manually-crafted features and leverage the advantage of end-to-end mechanism \cite{WangLS17, WangWCZL18, ChiangC19, HuangLLY18}.


Because mathematical expressions can be represented as binary trees, researchers have proposed seq2tree architecture that showed superior performance \cite{XieS19}. To enrich the question representations, Graph2Tree architecture was developed to build direct connections between numerals and entities in questions \cite{ZhangWLBWSL20, ShenJ20}. However, existing models mainly focus on the inference ability while the problem is often simply encoded as a sequence of words by one single vector. There is insufficient information available for decoding because the logical information hidden behind the structure among words is usually ignored. Take the structural relations between ``ratio'', ``transported goods'', ``rest'', ``is'' and ``3:7'' in Figure~\ref{fig:1} as example. According to the rule of semantic role labeling (SRL), this clause has ``is'', ``ratio'', and ``3:7'' as its \emph{root}, \emph{agent}, and \emph{patient} nodes, respectively \cite{SunSWW09}. In addition, ``transported goods'' and ``rest'' stand for the extent. Based on the semantic role labels, the model is expected to learn that ``3:7'' is a ``ratio''. Then, the model would be able to realize that the ``ratio'' describes the relationship between ''transported goods'' and ``rest''. Besides, learning the direct relations between numbers is important, because they will decide the order of decoding. For example, the model should prefer {7/(3+7)-(2/5)} more than {(2/5)-7/(3+7)}.

\begin{figure}[t]
    \centering
    \includegraphics[width=\linewidth]{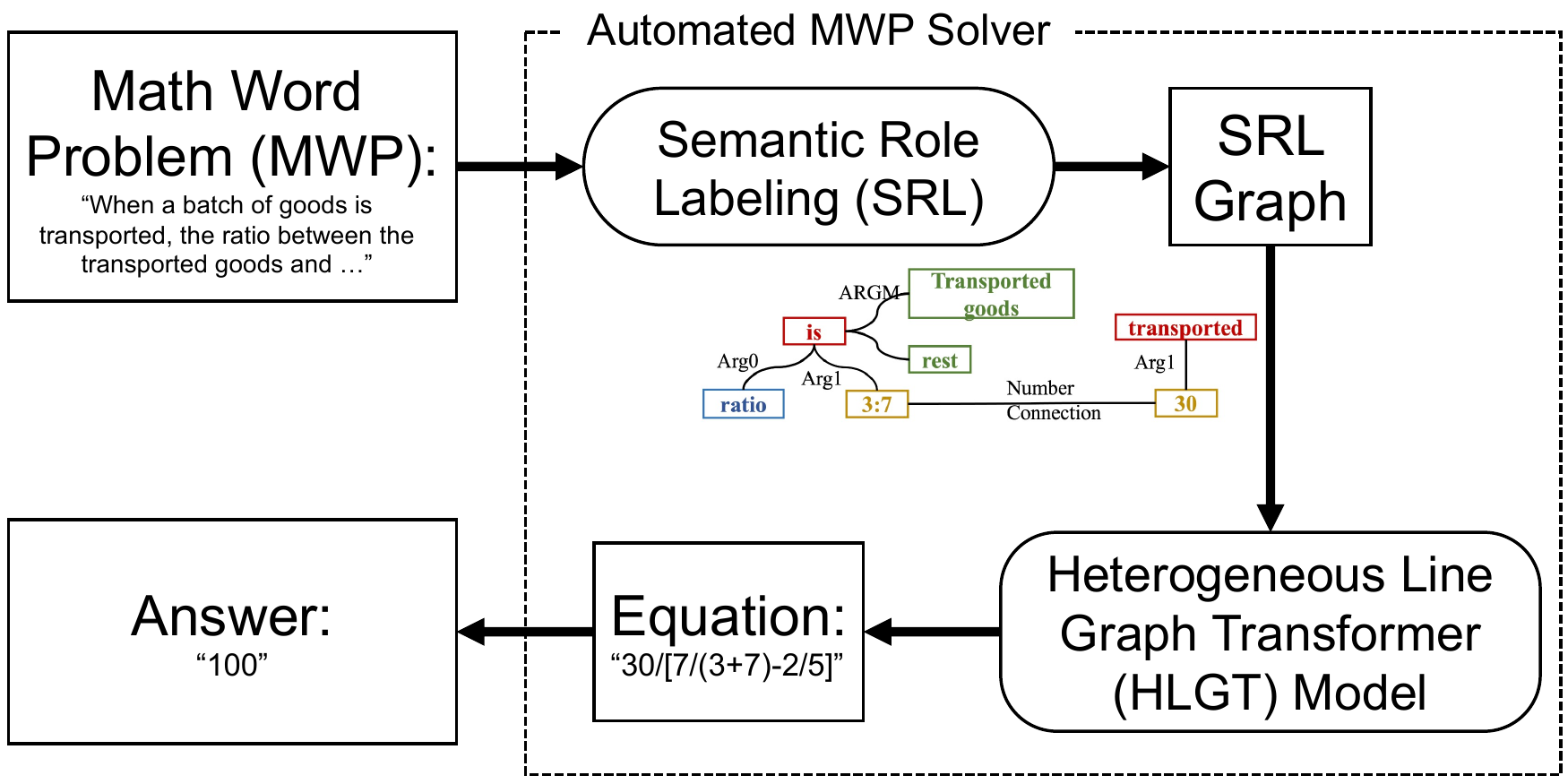}
    \vspace{-0.2in}
    \caption{Data forms and modules in the automated math word problem (MWP) solver system: The modules include a semantic role labeling (SRL) tool and the proposed heterogeneous line graph transformer (HLGT) model to produce equations from SRL graphs. An example of a MWP and SRL graph was given in Figure~\ref{fig:1}.}
    \label{fig:system}
    \vspace{-0.1in}
\end{figure}


After we tested with the existing solutions, we realized that unfortunately they were not able to achieve the desired performance due to the following issues. First, different types of connections in the input SRL graph were treated in the same manner. As a result, each node attended to neighborhoods equally \cite{ChenLLLZS20}, thus significant and unimportant relationships gave equivalent effects on the node representations. Second, node types and edge types were not considered in the model. The type information could provide useful prior knowledge for the model to learn to understand the complex relationships \cite{ChenXCXZSWQC20}. For example, in Figure~\ref{fig:1}, the number ```30'' is an integer representing the quantity of goods while ``2/5'' is a fraction indicating the proportion. Third, the connections across SRL subgraphs were ignored. For example, to understand the relationships between numbers, the meaning of each numerical token depends on other numerical tokens and provide supplementary details. e.g., ``2/5'' represents the proportion of the ``remaining goods'' to the ``original goods'' and ``30'' is the transported part. The implicit meta relation between ``2/5'' and ``30'' should be captured: ``original goods'' $\rightarrow$ ``3:7'' $\rightarrow$ ``30 tons'' $\rightarrow$ ``2/5''.

To address these issues, we propose a heterogeneous line graph transformer (HLGT) model to incorporate the structural information. First, we construct a heterogeneous graph according to the definition of semantic role labeling \cite{SunSWW09}. Different types of nodes and edges are extracted from the input objects, whose embedding vectors are used to capture the structural meaning. Given this graph, each node could gather structural dependency from neighborhoods and discover implicit relations automatically. Then, to have different considerations for different types of connections, we divide original graph into node-centered and edge-centered graphs, called Line Graph \cite{ChenLB19}, which explicitly considers the subgraph connections. Over the Line Graph, edge features can be updated like node features to preserve the context of meta relations. In the proposed model, we introduce a heterogeneous mechanism to incorporate the type information. It helps create and maintain the specific representation spaces for problems. In addition, we employ numerical comparison as an auxiliary task to improve the generality of the encoder which help identify the interactive relationship between the numbers.

We develop an automated MWP solver that integrates the HLGT model for producing equation expressions from SRL graphs. We apply and evaluate the MWP solver on a few real-world datasets such as Math23K \cite{WangWCZL18}, MAWPS \cite{Koncel-Kedziorski16}, and APE210K \cite{abs-2009-11506}. Experimental results demonstrate that HLGT achieves remarkable performance. Our major contributions are summarized as follows:
\begin{compactitem}
\item We transform the math word problems (MWPs) into heterogeneous line graph. Dual graph encourages question representation models to discover the implicit meta relationships between word tokens automatically.
\item We propose a novel model (HLGT) to enrich question representations. Type-specific parameters and edge dynamic embeddings are adopted to guide message aggregation from neighborhoods and capture the structure information.
\item We develop an automated MWP solver based on the HLGT model, and we conducted experiments to demonstrate the effectiveness of our model. Results show that the HLGT model achieves better performance than existing solutions.
\end{compactitem}

\section{Preliminaries}

In this section, we present the formal problem definition and introduce some basic knowledge of heterogeneous line graph.

Given a math word problem with a sequence of $n$ words $X=(x_1, x_2, x_3,...x_n)$, the goal is to generate a equivalent expression tree $Y=(y_1, y_2, y_3,...y_m)$. It consists of a set of variables $y_m \in \{o_1,o_2,o_3,..., o_i\}\cup\{c_1,c_2,c_3,...,c_j\}\cup\{q_1,q_2,q_3,...,q_k\}$, where $o, c$ and $q$ stand for operators (e.g., $+$, $-$, $\times$ and $/$), frequently used constants (e.g., 1, 100, 3.14, etc.) and the Numerical tokens existed in problem statements, respectively. Follow the settings in \cite{XieS19}, all the numerical tokens existed in problem $X$ are mapped to a special token ``NUM'' for we do not care the exact value of these tokens.

\begin{figure*}[t]
  \centering
  \includegraphics[width=\linewidth]{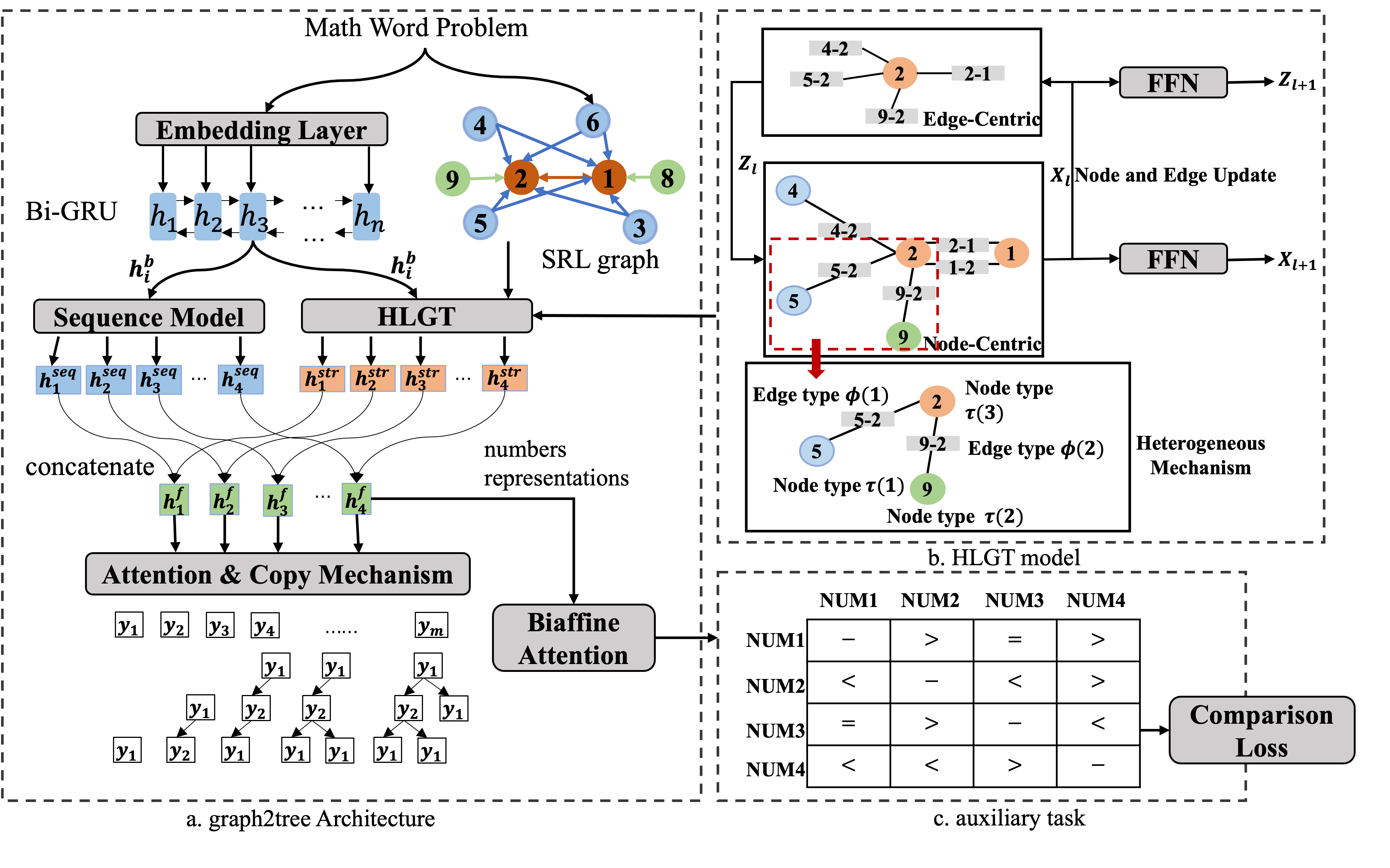}
  \vspace{-0.45in}
  \caption{The overall model architecture. Our model has multi-encoders and a tree-based decoder (a). To explicitly incorporate both sequential and structural information, we employ sequence and graph-structure encoders (b). The structure encoder takes the initial representation, heterogeneous SRL graph, and its line graph as inputs to update contextualized representations of nodes and edges. In this process, type and graph structure information are encoded into $h_i^{str}$. Simultaneously, the sequence encoder creates the sequential representation $h_i^{seq}$. We train a biaffine attention module on an auxiliary task (c) to compare numerical values with a comparison loss.}
  \label{fig:model}
  \vspace{-0.1in}
\end{figure*}

\begin{definition}[Heterogeneous Graph]
A heterogeneous graph \cite{2012Sun} is defined as a directed graph $G=\{\mathcal{V},\mathcal{E},\mathcal{A},\mathcal{R}\}$ with various types of nodes and edges, where $\mathcal{V}$ is a complete set of all nodes, $\mathcal{E}$ is a complete set of all edges. And each node $v\in\mathcal{V}$ and each edge $e\in\mathcal{E}$ are associated with their type mapping functions $\tau(v):V\rightarrow\mathcal{A}$ and $\phi(e):E\rightarrow\mathcal{R}$, respectively.
\end{definition}

\begin{definition}[Meta Relation]
For an edge $e=(s, t)$ linked from source node $s$ to target node $t$, its meta relation is denoted as <$\tau(s), \phi(e), \tau(t)$>. It describes a implicit relation from source node to target node. In this work, we defined 7 node types (i.e., entity, root, unit, rate, fraction, percentage, and other numbers.) and 7 edge types (see Table~\ref{tab:edgeandnode} for details) according to the rule of semantic role labeling \cite{SunSWW09} and numerical comparison results. In Table~\ref{tab:edgeandnode}, $e$, $r$, $u$, $q$ refers to entity, root, rate or unit, and numerical tokens in problem statements, respectively. \emph{BAE} and \emph{LES} are used to connect two number tokens, and representing $\geq$ and $<$ relation respectively. \emph{DT} is set for two numbers with different type.
\end{definition}

\begin{definition}[Line Graph]
Given an undirected graph $G_n = (V; E)$, its line graph $G_e = (V_L;E_L)$ encodes the directed edge adjacency structure of $G_n$. If $G_n$ is a node-centered graph, then the line graph $G_e$ could be viewed as edge-centered graph. For example, for edge $E(i\rightarrow j)$ and $E(j\rightarrow k)$ in graph $G_n$, they also act as two nodes $V_L(i,j)$ and $V_L(j,k)$ in line graph, and there is a directed edge $E_L(i,j\rightarrow j,k)$ between them. The vertices of $G_e$ consist of the ordered edges in $E$ (i.e., $V_L =
\{(i\rightarrow j):(i,j)\in E\}$), which indicates that $|V_L|=2|E|$. The edge set $E_L$ of $G_e$ is given by the
non-backtracking matrix $B\in \mathbb{R}^{2|E|\times 2|E|}$, which is defined as:
\begin{equation}
\label{eq1}
B_{(i\rightarrow j),(i'\rightarrow j')}=\left\{
\begin{aligned}
1 & , & if j=i' and j'!=i, \\
0 & , & otherwise.
\end{aligned}
\right.
\end{equation}
This matrix \cite{ChenLB19} enables the directed propagation of information on the line graph.
\end{definition}

\section{Proposed Model}

The proposed method is illustrated in Figure~\ref{fig:model}. It is inherited from the seq2tree structure and consists of multi-encoders and a tree-based decoder. Specifically, given a math word problem $X$, the word representation initialization module firstly generates initial representations for sequence encoder and structure encoder. Then, these two encoders work in parallel and output embedding vectors with different information. For word order information understanding, sequence encoder adopts a generalized recurrent neural model. For capturing the implicit and complex meta relations between word tokens, we design a heterogeneous line graph transformer (HLGT) module to discover the logical information and encode them into the word representation embedding vectors. In tree-based decoder, the model generates the expression tree $Y$ according to question representations via attention and copy mechanism.

\subsection{Words Representation Initialization}

We use a two-layer Bidirectional Gated Recurrent Unit (BiGRU) model \cite{ChoMGBBSB14} to encode the word-level representations. Formally, the BiGRU model takes the word sequence of problem statements $X=(x_1, x_2, x_3,...x_n)$ as input, and produces a sequence of hidden states $H = (h_1^x, h_2^x, h_3^x,...h_n^x)$ as output, where $\mathop{h_i^x}\limits ^{\rightarrow}$ and $\mathop{h_i^x}\limits ^{\leftarrow}$ represent forward hidden state vector and backward hidden state vector, respectively:
\begin{equation}
\label{eq2}
\left.
\begin{aligned}
& h_i^x = [\mathop{h_i^x}\limits ^{\rightarrow}; \mathop{h_i^x}\limits ^{\leftarrow}],\\
&\mathop{h_i^x}\limits ^{\rightarrow} = BiGRU(E(x_i), \mathop{h_{i-1}^x}\limits ^{\rightarrow}),\\
&\mathop{h_i^x}\limits ^{\leftarrow} = BiGRU(E(x_i), \mathop{h_{i-1}^x}\limits ^{\leftarrow}).\\
\end{aligned}
\right.
\end{equation}
Here, word embedding vectors $E(x_i)$ are obtained
via a wording embedding layer.
The representations $h_{i}^{x}$ has contained semantics of the word itself and incorporated contextual information, which serves as the basis of problem understanding for sequence and structure encoders:
\begin{equation}
\label{eq3}
h_i^b = \mathop{h_i^{x}}\limits ^{\rightarrow} + \mathop{h_i^{x}}\limits ^{\leftarrow}.
\end{equation}

\begin{table}
\caption{Checklist of edge types used in this work. $e$, $r$, $u$ and $q$ represent entity-, root- and unit- or rate- and numerical nodes, respectively. All the edge-types for node pair $(r, e)$ are symmetric.}
\label{tab:edgeandnode}
\vspace{-0.1in}
\begin{tabular}{c|c|c}
    \toprule
    Node pair&Edge type&Description\\
    \midrule
    \multirow{2}{*}{$(r, e)$} & ARG-0 & From \emph{root} to the \emph{agent}\\
                           & ARG-1 & From \emph{root} to the \emph{patient}\\
                           & ARG-M & From \emph{root} to the \emph{auxiliary roles}\\
    \hline 
    $(e, e)$& MOD & Relationship between the \emph{entities}\\
    \hline 
    $(u, q)$ & MOD & The \emph{rate} or \emph{unit} to the related entity\\
    \hline 
    \multirow{2}{*}{$(q,q)$} & BAE & Better and equal\\
                           & LES & Less\\
                           & DT & Different types\\
  \bottomrule
\end{tabular}
\vspace{-0.1in}
\end{table}

\subsection{Heterogeneous Graph Construction}

We refer all number tokens $X_q$, nouns $X_n$, units/rates $X_u$ and predicate verbs(root) $X_r$ as possible nodes of node-centered graph. Next, we define a SRL graph as a subset of nodes in the problem. Specifically, we firstly use string matching to extract all Numbers as numerical nodes. Then, we retrieve all phrases and their labeled semantic roles through natural language processing toolkit \cite{hanlp2, che2020n}. Among the phrases, the token labeled as \emph{root} is regarded as root-node. Next, we identify all the nouns in extracted phrase as entity nodes based on the Part-of-Speech Tagging (PoS) result. At last, if a number token is immediately followed by a unit or rate word, this word will be recognized as a unit or rate-node. 

We construct 7 kinds of edges to provide different connections between nodes. Formally, the edges $\mathcal{E}$ label the relationship between the nodes, which correspond to the following four situations.   
\begin{compactitem}
\item \emph{Edge between root and entity:} A directed edge exists between root-node and entity-node in the same SRL section, which is defined according to the annotations of SRL.
\item \emph{Edge from entity to entity:} For non-root phrases with more than one node, we select one of the them as the contact node and link it to the root node. If there is a numerical node in the phrase, then this node will be the contact node, otherwise the last occurring entity node will be selected. This edge is defined to connect non-contact and contact nodes in the same role phrase.
\item \emph{Edge from unit/rate to number:} For each number and its related unit or rate, a directed edge \emph{MOD} is added. Because the unit- or rate-node may contain specific algebraic calculation information (such as * 10 are needed for the conversion of centimeters to decimeters).
\item \emph{Edge from number to number:} For node pairs with the same type, an edge containing quantity comparison information would be added (i.e., better and equal \emph{BAE}, less \emph{LES}). For different types of numbers, it is meaningless to compare them, so we use different type \emph{DT} edge to connect them. The edges in this situation cluster the same type numbers together and conduct numerical comparison. This edge also serves the function of connecting subgraphs.
\end{compactitem}

\subsection{Heterogeneous Line Graph Transformer}

Here, we present the details of HLGT module, which combines the advantages of line graph \cite{ChenLB19} and heterogeneous multi-head attention mechanism \cite{HuDWS20}. Based on the heterogeneous graph $G$, HLGT enriches question representations, which collects the meta relation information through multiple iterations of message passing between neighborhoods. For instance in Figure~\ref{fig:1}, the goal is to calculate how many tons of this batch of goods. The HLGT model is expected to discover the meta path between \emph{30} and \emph{origin good} through aggregating information from neighbors.

\paragraph{\textbf{Origin Graph}} Given the node-centered graph $G_n$, the output representation $h_{i}^{l}$ of the l-th iteration is computed by equation~\ref{eq4}. 
\begin{equation}
\label{eq4}
\left.
\begin{aligned}
&\tilde{a_{ij}} = (h_{i}W_q^{\tau(x_i)})(h_{j}W_k^{\tau(x_j)})\odot\psi(z_{ji}),\\
&a_{ij}=softmax(\tilde{a_{ij}}/\sqrt{d/M}),\\
&\tilde{h_{i}}=||_{m=1}^M\sum\limits_{v_j\in\mathcal{N}_i^n}a_{ij}^m(h_{j}W_v^{\tau(x_j)}\odot\varphi(z_{ji})),\\
&h_{i}^{l+1}=LayerNorm(\delta^{\tau(x_i)}h_{i}+(1-\delta^{\tau(x_i)})\tilde{h_{i}}U_o^{\tau(x_i)}),
\end{aligned}
\right.
\end{equation}
where $||$ represents vector concatenation, matrices $W_q^{\tau(x_i)}, W_k^{\tau(x_j)}$, $W_v^{\tau(x_j)}\in\mathbb{R}^{d\times d/M}$, $\psi,\varphi$ and  $U_o^{\tau(x_i)}\in\mathbb{R}^{d\times d}$ are trainable parameters, M is the number of heads and $LayerNorm()$ denotes a normalization layer.

Firstly, $\tau(x_i)$-type source node representation and $\tau(x_j)$-type target node representation are projected into type-specific Key vector $W_k^{\tau(x_j)}$ and Query vector $W_q^{\tau(x_i)}$, respectively. Then, We use heterogeneous mutual attention mechanism to calculate multi-head attention $a_{ij}$. 
To maximize parameter sharing while discovering the meta relations between different node pairs, we use $\psi$ vectors to project edge-features $z_{ji}$, and then multiply the result with source and target node projection. Next, in message passing part, we also incorporate the dynamic edge features to alleviate the distribution features of implicit meta relations. We replace all edge-type parameters matrix and use dynamic edge representation calculated by the edge-centered module. This way is more flexible than the fixed size parameter matrix to capture the relationship between numbers or between numbers and entities. At last, the node representation needs to be updated over the heterogeneous multi-head attention and type-specific passing message. A type coefficient $\delta^{\tau(x_i)}$ is adopted to decide how much information is retained from the original representation and how much information aggregated from neighbor nodes. Besides, the passing message $\tilde{h_{i}}$ is mapped to type-specific message output vector $U_o^{\tau(x_i)}$, so that to get the different type distribution.

\paragraph{\textbf{Line Graph}} Given edge-centered graph $G_e$, the updated node representation $z_{l+1}$ could be calculated with multi-head attention mechanism:

\begin{equation}
\label{eq5}
\left.
\begin{aligned}
&z_i^0=E_e(edge_i),\\
&b_{ij}=\frac{exp(LeakeyReLU([S(z_i)||S(z_j)||\rho(h_{ji})]))}{\sum_{k\in\mathcal{N}_i}exp(LeakeyReLU([S(z_i)||S(z_k)||\rho(h_{ki})]))},\\
&z_{i}=||_{m=1}^M\sum\limits_{v_j\in\mathcal{N}_i^e}b_{ij}^m(S_v[z_{j}||\rho(h_{ji})]),\\
&z_{i}^{l+1}=FFN(z_{i}),
\end{aligned}
\right.
\end{equation}

where we use one embedding layer $E_e$ to obtain the initialized representation $z_i^0$ of node in edge-centered graph. In equation ~\ref{eq5}, $\rho(h_{ji})$ returns the feature vector of relation $h_{ji}$ in $G_e$, which is retrieved from node representations $h_{i}$ in $G_n$. $S, S_v\in\mathbb{R}^{d\times d}$ are trainable parameters, d is the dimensions of $z_{i}$. $FFN$ denotes a feed forward network layer.

Considering the directed graph used in this work, we pay more attention to the updating of the target node. As a result, mixed static-dynamic embedding is adopted to provide different embedding for source and target node in line graph. if $e_{ij}$ is a target node in $G_e$, $z_{ij}$ returns the node embeddings in line graph. Otherwise, $z_{ij}$ directly retrieves the vector from the initial edge embedding matrix $E_e$.

Through the whole HLGT model, we could use node type $\tau(s)$ to parameterize the linear vector matrix separately. Type-specific parameters help the model to learn implicit relationship between numbers and other nodes. At the same time, over line graph concept, the static edge features are transformed into dynamic features which keep the independence of edges in different graph structure. Compared with existing models, HLGT's dynamic embedding can better leverage the heterogeneous graph structure to enrich question representations. On one hand, nodes with few neighbors can further obtain non-local information through edge features. On the other hand, in addition to the node information, structural information is also used to infer implicit meta relations. 

\subsection{Tree-structured Decoder}

We use Goal-driven Tree Structure (GTS) decoder \cite{XieS19} as our decoder module. Firstly, the final problem representations are obtained by concatenating the sequence hidden states $h_{i}^{seq}$ and the structure hidden states $h_{i}^{str}$:
\begin{equation}
\label{eq6}
\begin{aligned}
&h_i^{seq} = BiGRU(h_i^{b}, h_{i-1}^{seq}),\\
&h_i^{str} = h_i^{L},\\
&h_i^f = [h_i^{seq}: h_i^{str}],\\
\end{aligned}
\end{equation}
where $h_i^{b}$ is the output of word representation initialization module, $h_i^{L}$ is the output of the last layer of HLGT module.
Then, the tree structured decoder takes the final problem representations $h_i^f$ as input and generates the target expression from top to bottom. Specifically, decoder will predict whether the current token should be decomposed further according to the context vector and goal vector (see \cite{XieS19} for more details). If the predicted token is a number in problem statements or constant quantity, the current goal is realized directly; otherwise (i.e., the predicted token is an operator), two new sub-goal vectors (one for left sub-goal and the other for the right) will be generated. The embedding $e(y|X)$ For predicted token could be depicted as:
\begin{equation}
\label{eq7}
e(y|X)=\left\{
\begin{aligned}
& e(y|op), if y\in V_{p,op}, \\
& e(y|cons), if y\in V_{p,cons}, \\
& H_q^L, if y\in V_{p,q}, \\
\end{aligned}
\right.
\end{equation}
where if $y$ is an operator or a constant, its embedding should be retrieved from a trainable parameter matrix directly. If $y$ is a number in a problem statement, its embedding is copied from the final representations of problem statements. Due to the decomposition process is implemented as the pre-order operation during the traversal, when predicting sub-goal token, the model will first process the left sub-goal. When it comes to the right sub-goal, the construction of its left sub-tree must have been completed. Then, the embedding of all left sub-tree nodes will be recursively encoded to the root node of the left sub-tree and used in predicting the target of the right sub-tree.

\subsection{Numerical Comparison}

A generalized encoder for MWP should be able to compare the value of numbers. When we consider how to calculate the difference of two numbers $q_1$ and $q_2$, the comparative relationship between them can help decide the order they appear in the expression (e.g. $q_1-q_2$ or $q_2-q_1$). To introduce this inductive bias, we design an auxiliary task to compare the  relationship between two numbers.
We retrieve all the numerical token representation vectors $h^{f}_{q}$ from the final question embeddings. Then, we use a biaffine \cite{DozatM17} binary classifier to determine the comparative relationship between the number pair ($q_1, q_2$).
\begin{equation}
\label{eq8}
\begin{aligned}
&Biaffine(q_1; q_2) = h^{f}_{x_{q1}}U_sh^{f, T}_{x_{q2}} + [h^{b}_{x_{q1}};h^{b}_{x_{q2}}]W_s+b_s,\\
&p^{com}(q_1; q_2) = \sigma(Biaffine(q_1; q_2)).
\end{aligned}
\end{equation}
Then the ground truth label $y^{com}(q_1; q_2)$ is determined by the relationship between the number pair (e.g., $>=$ is 1 and $<$ is 0). And the training object can be formulated as
\begin{equation}
\label{eq9}
\mathcal{L}^{com} = 1/K^2\sum_{i=1}^{K}\sum_{j=1}^{K}p^{com}(q_i; q_j),
\end{equation}
where K is the number of numerical tokens in problem.

\subsection{Training Method}

Given a batch of question statements ${X}$ and its corresponding ground truth mathematical expression ${Y^g}$, the training object $\mathcal{L}^q$ could be formulated as the sum of negative log-likeihoods of probabilities for predicting expression $Y^p$. And the final loss function $\mathcal{L}$ is the weighted sum of $\mathcal{L}^q$ and  $\mathcal{L}^{com}$:
\begin{equation}
\label{eq10}
\begin{aligned}
&Pr(Y^p|Y^g,X)=\prod_{t=1}^{m} Pr(y_t|Y^g_{t-1},G,X),\\
&\mathcal{L}^q=\sum_{(X,Y)\in \mathcal{D}}-\log Pr(Y^p|Y^g,X),\\
&\mathcal{L} = \mathcal{L}^q + \beta^{com}\mathcal{L}^{com},
\end{aligned}
\end{equation}
where $y_t$ is the t-th token predicted by the model and $G$ represents the graph input.

\section{Experiments}

In this section, we evaluate the proposed model in different settings. Ablation study and analysis are also conducted to demonstrate the effectiveness of its components.

\subsection{Experimental Settings}

\paragraph{\textbf{Datasets}}
Math23K \cite{WangLS17} is a Chinese dataset that contains 23,162 elementary school level MWPs and corresponding equation expressions and answers. 
MAWPS \cite{Koncel-Kedziorski16} is an English dataset with 2,373 math word problems. APE210k \cite{abs-2009-11506} is a large-scale dataset with 200,488 training data, 5000 test data and 5,000 valid data. We pre-processed APE210k to ensure that its mathematical expression can be executed, and the execution result is consistent with the answer. After processing, 162,081 training data, 4,046 test data and 4,057 valid data were kept. 

\paragraph{\textbf{Baseline methods}}
We compared our method with various baselines and state-of-the-art model: \textbf{T-RNN} \cite{WangWCZL18} uses equation normalization method to normalize the duplicated equations and applies seq2seq structure to generate equation expression. \textbf{GROUP-ATTENTION} \cite{LiWZWDZ19} designs a multi-head attention module with different meanings referring to the transformer mechanism. \textbf{GTS} \cite{XieS19} decomposes the problem into a binary tree composed of sub-goals. \textbf{Graph2Tree} \cite{ZhangWLBWSL20} 
adopts single layer of graph encoder to enrich the number features. \textbf{MEMD} \cite{ShenJ20} proposes a model with multi-encoders and multi-decoders, and designs a new graph form for math word problem. \textbf{FECA} \cite{abs-2009-11506} enhances seq2seq model with external features and copy mechanism. \textbf{NUMS2T} \cite{WuZWH20}
treats all numbers with explicit numerical values and adds external knowledge into encoder model.

\paragraph{\textbf{Implementation}}
We utilize PyTorch \cite{PaszkeGMLBCKLGA19} to implement our method. Specifically, question is first tokenized by \emph{Hanlp} \cite{hanlp2} for English datasets and \emph{LTP} \cite{che2020n} for Chinese datasets. For the input representation, We initialize the word embedding with the pretrained 300-dimension word vectors. The number of heads M in HLGT is set to 4, and the number of HLGT layer is 2. The hidden size is 512 and the batch size is 64. For optimizer, we use Adam \cite{KingmaB14} optimizer to minimize loss and set the beginning learning rate as 1e-3. In the training process, the learning rate is halved every 20 epochs for Math23K and 10 for APE210K/MAWPS. The numerical comparison loss coefficient $\beta^{com}$ is set to 0.1. Models are trained in 80 epochs for Math23K, 50 epochs for APE210K, and 30 epochs for MAWPS. During testing, the beam size is set as 5.


\subsection{Main Results}

\begin{table}
  \caption{Answer accuracy on three test datasets.}
  \label{tab:result}
  \vspace{-0.1in}
  \begin{tabular}{l|ccc}
    \toprule
      &Math23K& MAWPS & APE210K\\
    \midrule
    {T-RNN\cite{WangWCZL18}} & 66.9& 66.8 & -\\
    {GATT \cite{LiWZWDZ19}} & 66.9& 76.1 & -\\
    {GTS \cite{XieS19}} & 75.6 & 82.6 & 67.7\\
    {G2Tree \cite{ZhangWLBWSL20}} & 77.4 & 83.7 & 68.2\\
    {MEMD \cite{ShenJ20}} & 78.4 & - & -\\
    {FECA \cite{abs-2009-11506}} & 77.5 & - & 70.2\\
    {NUMS2T \cite{WuZWH20}} &78.1& 83.8 & 70.5\\
    {HLGT (ours)} & \textbf{79.0} & \textbf{84.8} & 69.8\\
  \bottomrule
\end{tabular}
\vspace{-0.1in}
\end{table}

The main results are provided in
Table~\ref{tab:result}. Our method achieves excellent expression accuracy of $79.0\%$ for Math23K, $84.8\%$ for MAWPS and $69.8\%$ for APE210K.

From the experimental results: 1) Performance of the models that further process natural language (i.e., Transforming natural language into graph (Graph2Tree, MEMD, ours), introducing external knowledge and dealing numerical token with explicit Values (NUMS2T)) is better compared with the models that simply treat natural language (FECA, GTS) as sequence. This indicates that there is indeed a gap between natural language and machine-readable forms. 2) In addition, Graph2Tree outperforms GTS model, which proves that graph inputs promote the model to learn implicit structure information and improve the model accuracy. 3) Compared with NUMS2T, Our model still shows advantages on Math23K and MAWPS datasets. NUMS2T uses the external knowledge base to extract nodes and edges from problem statements and construct a input graph. However, our model learns hidden meta relationships from SRL graphs automatically. This illustrates that HLGT model could learn meta relations between words without external knowledge.

Although APE210k is the largest dataset, they contains either some unlabeled problems or unexecutable equation expressions. It requires to clean the dataset before training. Due to different processing methods, FECA, NUMS2T and HLGT retain valid set with different number of items.

\subsection{Ablation Study}

In this section, we investigate the contribution and effectiveness of each components.

\paragraph{\textbf{Effect of Graph Construction Method}}
We firstly tested the proposed model on three different heterogeneous graphs:
\begin{compactitem}
\item {Parse Graph + Numerical Comparison Graph \cite{ShenJ20}:} If a word pair has dependency connection, there is an undirected edge between these two words. In addition, for two number tokens $q_i,q_j$, there are two directed edges that represent the quantity comparison relationship in both direction.
\item {Quantity Cell Graph + Quantity Comparison Graph \cite{ZhangWLBWSL20}:} This construction uses pre-processing tools to extract verbs, nouns, adjectives and units related to the numbers. Undirected edges are added to the extracted words and the corresponding numerals. Quantity comparison graph is similar to numerical comparison graph, where only a single directed edge stands for the $>$ relationship are retained.
\end{compactitem}

\begin{table}
  \caption{Comparison among graph construction methods (NC: Numerical Comparison).}
  \label{tab:graph}
  \vspace{-0.1in}
  \begin{tabular}{l|cc}
    \toprule
     & Math23K& MAWPS\\
    \midrule
    Parse graph + NC \cite{ShenJ20} & 76.5& 82.9\\
    Quantity cell + NC \cite{ZhangWLBWSL20} & 77.4 & 83.7\\
    SRL + NC + line graph & \textbf{79.0} & \textbf{84.8}\\
  \bottomrule
\end{tabular}
\vspace{-0.1in}
\end{table}

As shown in Table~\ref{tab:graph}, the expression accuracy of the heterogeneous line graph is obviously better than the other two graph construction methods. It may be caused by the following reasons: a) The first construction method lacks the direct connection between numeral and other word tokens, so the model can not learn the implicit meta relationships between them. b) The second graph construction method only connects numerals with their related words, which leads to the lack of the global context for number representation.

\paragraph{\textbf{Effect of Different Components}}
We verified the overall performance improvement brought by each module. Here, ``HLGT w/o node type'' means that the node type matrix were removed from the model. ``HLGT w/o line graph'' means that we delete line graph edge updating module. ``HLGT w/o auxiliary task'' means that the final loss function excludes Numerical comparison loss. In addition, we compare HLGT with other graph neural network modules that are good at dealing with the relationship between nodes: 1) \emph{Relational-aware schema graph attention network (RGAT)} \cite{WangSLPR20} 
adds edge feature to the formula as an additional term when calculating self-attention; 2) \emph{Heterogeneous Graph Transformer (HGT)} \cite{HuDWS20} designs node- and edge-type dependent parameters for model graph heterogeneity.

From Table ~\ref{tab:ablation}, we could found that: 1) if
node-type or line graph are removed, the performance decreases. This consolidates our two motivations that by exploiting the structure of problems and subgraph connections, the model can capturing implicit meta relationships between tokens. 2) Without auxiliary task, the expression accuracy are reduced to 78.5\% and 84.0\% for Math23K and MAWPS datasets respectively, which may be caused by the lack of the ability to compare the numerals. 3) The performance of HGT and RGAT is relatively weak without dynamic edge features captured by line graph module. This verifies that the line graph is helpful for the model to learn the implicit meta relationships. 4) The model with node-type (HLGT, HGT) information performs better than that without node-type (RGAT). It consolidates that node type can be used as a prior knowledge to help the model distinguish different types of nodes.

\begin{table}
  \caption{Effectiveness of different model components.}
  \label{tab:ablation}
  \vspace{-0.1in}
  \begin{tabular}{lcc}
    \toprule
     & Math23K& MAWPS\\
    \midrule
    HLGT w/o node type & 78.2 & 83.4\\
    HLGT w/o line graph & 77.8 & 83.2\\
    HLGT w/o auxiliary task & 78.5 & 84.0\\
    \hline
    RGAT\cite{WangSLPR20} & 77.0 & 82.9\\
    HGT\cite{HuDWS20} & 77.4 & 82.7\\
    \hline
    HLGT & 79.0 & 84.8\\
    \bottomrule
\end{tabular}
\vspace{-0.1in}
\end{table}

\paragraph{\textbf{Effect of Problem Complexity}} 
Table~\ref{tab:length} shows how well HLGT model performs with the increasing question complexity as compared to best-performing NUMS2T. Follow the experiment in \cite{WuZWH20}, we choose the number of numerals to distinguish questions difficulty. The result demonstrates that our proposed HLGT outperforms the baseline at all levels of difficulty. This is because HLGT could learn the meta relations in problems automatically, which could better understand the structure information between word tokens. In addition, HLGT brings more enhancements in complex cases. This may be due to the subgraph connectivity in HLGT model, which provides supplementary information for the inference. 

\begin{table}
\caption{Model performance on different difficulty of problems. Num.: numbers of numerical tokens in the problem.}
\label{tab:length}
\vspace{-0.1in}
\begin{tabular}{r|ccc}
    \toprule
    \multicolumn{4}{c}{Math23K}\\
    \hline
    Num. & Proportion. & NUMS2T & HLGT\\
    \midrule
    <=1 & 2.8\% & 83.0\%& 84.5\% (+1.5\%)\\
    2 & 36.2\% & 85.1\%& 85.3\% (+0.2\%)\\
    3 & 45.3\% & 78.4\% & 79.2\% (+1.8\%)\\
    4 & 12.3\% & 60.6\% & 67.0\% (+6.4\%)\\
    5 & 2.0\% & 54.9\% & 55.3\% (+0.4\%)\\
    6 & 0.7\% & 46.7\% & 56.2\% (+9.5\%)\\
    >=7 & 0.8\% & 37.5\% & 52.6\% (+15.1\%)\\
  \bottomrule
\end{tabular}
\vspace{-0.1in}
\end{table}

\begin{figure}[t]
  \centering
  \includegraphics[width=\linewidth]{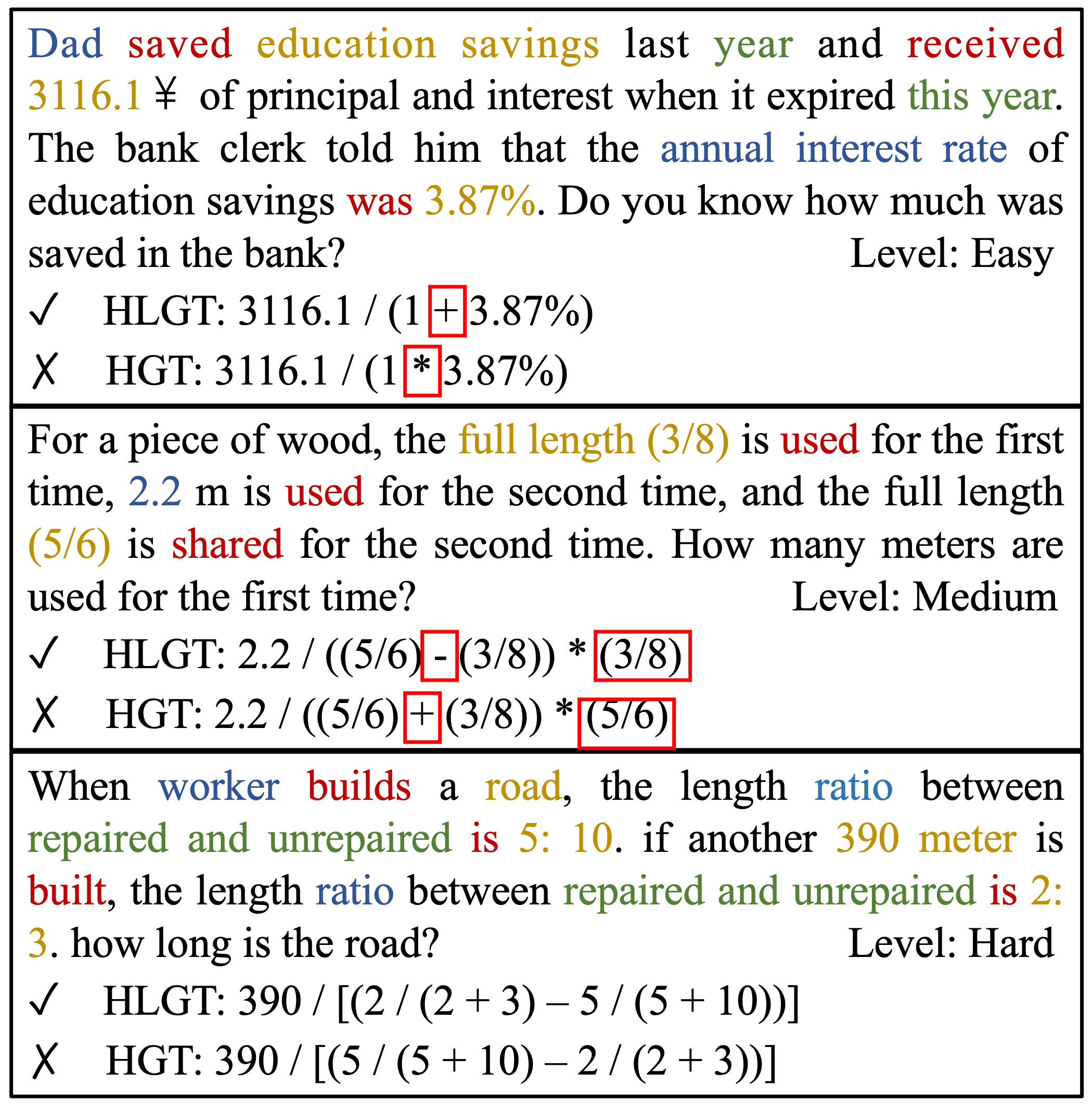}
  \vspace{-0.3in}
  \caption{Case study: Three examples of generated expression by HGT\cite{HuDWS20} and the proposed HLGT model.}
  \label{fig:3}
  \vspace{-0.1in}
\end{figure}

\subsection{Case Study}

To further evaluate the HLGT's advantage of capturing the meta relations between numbers, we conduct a case study on three samples. And we select HGT model as our control system without line graph. From Figure~\ref{fig:3}, we found that HLGT performs better than baseline system, especially on examples with multi-type of numbers and complex meta path. In case 1, \emph{3116.1} is the sum of principal and interest after one year, so the operator between \emph{1} and \emph{3.87$\%$} should be \emph{+}. However, HGT fails to capture the meta-relations between \emph{3116.1} and \emph{3.87$\%$}. For the second case, HLGT could discover that \emph{2.2} is the reason for the difference between \emph{(5/6)} and \emph{(3/8)} from meta path <$2.2, second~time, (5/6)$> and <$first~time, (3/8)$>. However, HGT cannot find the relationship between \emph{(5/6)} and \emph{(3/8)} which leads to the error. For last case, this MWP requires models to learn that the numbers in the question has similar context. So the key point is to learn that \emph{2:3} is the ratio after another 390 meter was repaired, while 5:10 is the ratio before the repair.

\subsection{Visualize Graph Attention}

\begin{figure}[t]
  \centering
  \includegraphics[width=\linewidth]{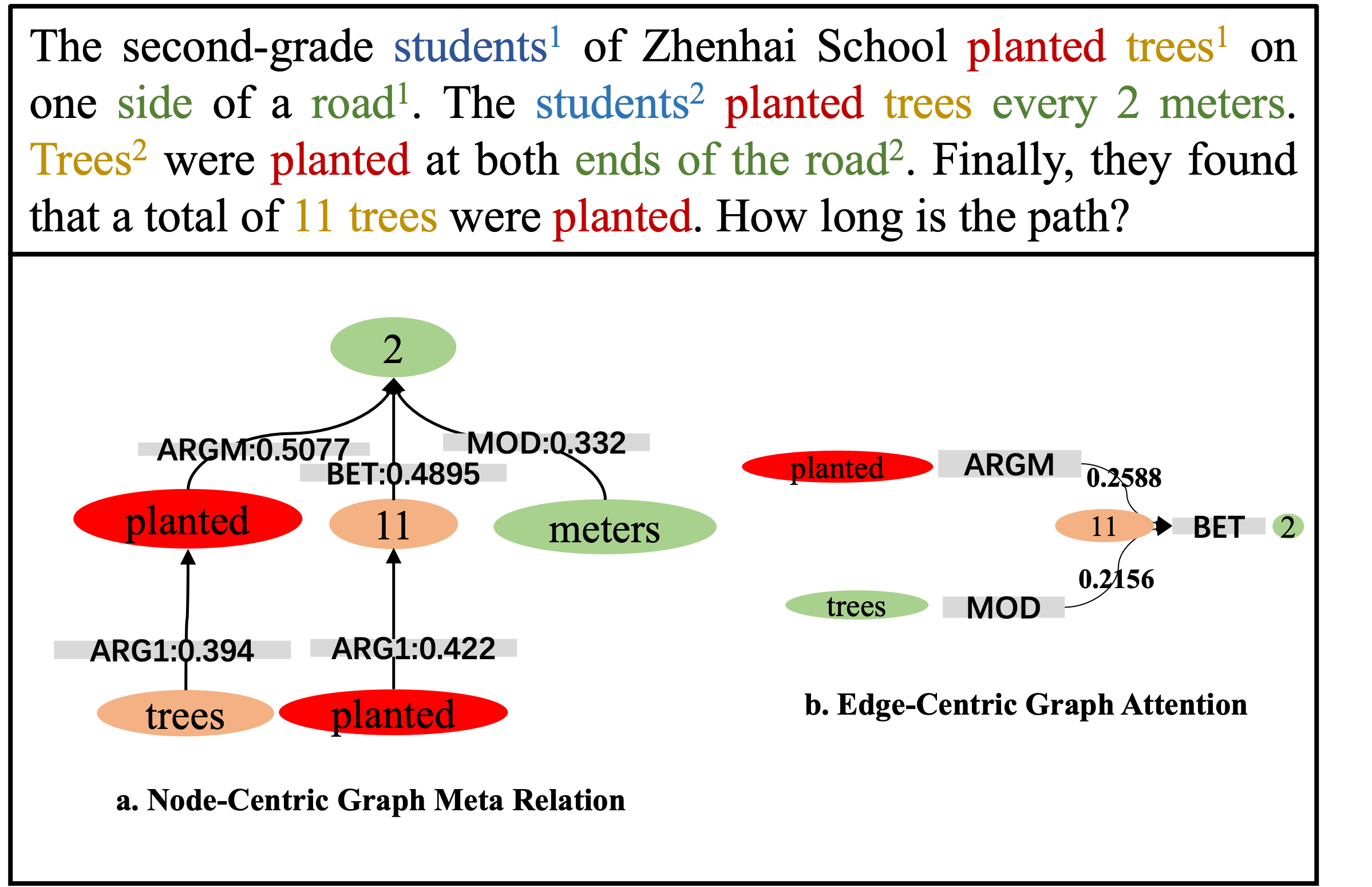}
  \vspace{-0.2in}
  \caption{Visualization of meta-relation attention that is trained on node-centric and edge-centric graphs.}
  \label{fig:4}
  \vspace{-0.1in}
\end{figure}

To prove that the HLGT model can learn the implicit meta relationship between different nodes automatically, we pick the neighbors that have the largest attention value and plot the meta relation attention in Figure~\ref{fig:4}. In this example, we choose number \emph{2} as the target. Among all meta relations, <trees, \emph{ARG1}, planted, \emph{ARGM}, 2>, <planted, \emph{ARG1}, 11, \emph{BET}, 2> are the two most important meta relation tracks. At the same time, edge \emph{BET} aggregates information from neighbor edges in different proportions in line graph module. This proves that HLGT model can automatically learn the meta-relationship between neighbor nodes and then decide how much information should be aggregated from others.

\section{Related Work}

\subsection{Math Word Problem}

MWP solutions could be categorized into three stages. In the first pioneering stage, The idea of rule-based matching is adopted \cite{Slagle65, MukherjeeG08}, in which manually-crafted rules plays an important role in the whole system. As a result, these solvers rely heavily on the pre-definition of mathematical expressions template.
In the second stage, semantic parsing \cite{ShiWLLR15, Koncel-Kedziorski15, RoyR15, GoldwasserR14, KwiatkowskiCAZ13} attracted a lot of research interest. This method focuses on feature engineering, that is, in addition to the original input, the problem statements are transformed into structured input to help the model perform better.
Recently, with the increase of training items number, the deep learning method has achieved absolute advantages in Math Word Problem \cite{WangLS17}.
Since the mathematical expression can be transformed into a symmetric binary tree, The idea of tree-based approaches was proved better performance among deep learning based methods. {Seq2SeqET} \cite{WangWCZL18} use seq2seq model to directly generate the equivalent expression tree. In addition, equality normalization is adopted to shrink the template space. {T-RNN} \cite{WangWCZL18} has similar ideas with {Seq2SeqET} while further improves the method of equality normalization. {GTS} \cite{XieS19} replaces the sequence decoding model with goal driven decoding, which decomposes the problem into a binary tree composed of sub-goals. 

In recent years, the advantages of graph neural network in processing graph structure data gradually appear. And it guides another branch line of MWP: the combination of semantic parsing and seq2seq method. This kind of method usually extracts the key information from problems to form a new graph, and add a graph neural network module on the encoding side. {Graph2Tree} \cite{ZhangWLBWSL20} 
designs a quantity-centered graph for problem statements and adopts a single layer of graph encoder. {MEMD} \cite{ShenJ20} proposes a model with multi-encoders and multi-decoders, and designs a different graph form with {Graph2Tree}. {NUMS2T} \cite{WuZWH20}
treats all numbers with explicit numerical values and adds external knowledge graph into encoder model. 

\subsection{Heterogeneous Graph Neural Networks}

To mine relationship between nodes in a heterogeneous graph with multiple types of edges and nodes \cite{2012Sun}, models should have the ability to handle node- and edge-type information. To maintain node-type spaces, \cite{ZhangSHSC19} integrates the node-type into the encoder. {RGCN} \cite{SchlichtkrullKB18} and {RGAT} \cite{WangSLPR20} design edge-type specific linear projection for attention mechanism. {HGT} \cite{HuDWS20} fully utilizes the heterogeneity of graphs with node- and edge-type specific parameters matrix. Graph structure is another key point for heterogeneous graph. To address static edge embedding for different graph structure, {GTN} \cite{CaiL20} use sequence model to generate different edge embeddings in different graph structure. Line Graph was adopted in \cite{ZhaoCCCZY20} to effectively explore the edge relations in different graph structure. 

\section{Conclusions}

In this paper, we designed a heterogeneous line graph transformer model to enrich the question representation for automated math word problem solver. To narrow the gap between the problem statements and the machine understandable form, we constructed a heterogeneous graph, over which model could learn meta-relations more directly than unprocessed natural language. We used line graph to discover the implicit meta relations paths between tokens. To help the model distinguish the features contained in different types of nodes, node type-specific linear projection was used in attention calculation. Comprehensive experiments on MWP datasets and showed that the proposed HLGT model captured meta relations in problem statements and achieves good performance.


\bibliographystyle{ACM-Reference-Format}
\bibliography{sample-base}

\end{document}